\documentclass[sigconf]{acmart}
\usepackage{amsmath,graphicx,hyperref}
\usepackage{booktabs}

\usepackage{amssymb}
\usepackage{multirow}
\usepackage{setspace}
\usepackage{subcaption} 
\usepackage{color}
\usepackage{xcolor}
\usepackage[table]{xcolor}
\usepackage[dvipsnames]{xcolor} 

\AtBeginDocument{%
  }

\setcopyright{acmlicensed}
\copyrightyear{2026}
\acmYear{2026}
\acmConference[ICMR '26]{Proceedings of the 2025 International Conference on Multimedia Retrieval}{June 16-June 19, 2026}{Amsterdam, Netherlands}
\acmISBN{978-1-4503-XXXX-X/2018/06}

\begin{document}

\title[DF-LLaVA: Unlocking MLLMs for Synthetic Image Detection]{DF-LLaVA: Unlocking MLLMs for Synthetic Image Detection via Knowledge Injection and Conflict-Driven Self-Reflection}



\author{Zhuokang Shen}
\affiliation{%
  \institution{East China Normal University}
  \city{Shanghai}
  \country{China}
}

\author{Kaisen Zhang}
\affiliation{%
  \institution{East China Normal University}
  \city{Shanghai}
  \country{China}
}

\author{Bohan Jia}
\affiliation{%
  \institution{East China Normal University}
  \city{Shanghai}
  \country{China}
}

\author{Heming Jia}
\authornote{Corresponding author.}
\email{jiaheming@fjsmu.edu.cn}
\affiliation{%
  \institution{Sanming University}
  \city{Fujian}
  \country{China}
}

\author{Yuan Fang}
\affiliation{%
  \institution{East China Normal University}
  \city{Shanghai}
  \country{China}
}
\affiliation{%
  \institution{The 27th Research Institute of CETC}
  \city{Zhengzhou}
  \country{China}
}

\author{Zhou Yu}
\affiliation{%
  \institution{East China Normal University}
  \city{Shanghai}
  \country{China}
}

\author{Shaohui Lin}
\authornote{Corresponding author.}
\email{shlin@cs.ecnu.edu.cn}
\affiliation{%
  \institution{East China Normal University}
  \city{Shanghai}
  \country{China}
}
\affiliation{%
  \institution{Sanming University}
  \city{Fujian}
  \country{China}
}

\renewcommand{\shortauthors}{Shen et al.}

\begin{abstract}
With the increasing prevalence of synthetic images, evaluating image authenticity and locating forgeries accurately while maintaining human interpretability remains a challenging task. 
Existing detection models primarily focus on simple authenticity classification, ultimately providing only a forgery probability or binary judgment, which offers limited explanatory insights into image authenticity. 
Moreover, while MLLM-based detection methods can provide more interpretable results, they still lag behind expert models in terms of pure authenticity classification accuracy.
To address this, we propose \textbf{DF-LLaVA}, a novel and effective framework that \textit{unlocks the intrinsic discrimination potential of MLLMs.}
Our approach first mines latent knowledge from the MLLM itself and then injects it into the model via fine-tuning. 
During inference, conflict signals arising from the model’s predictions activate a self-reflection process, leading to the final refined responses.
This framework allows LLaVA to achieve outstanding detection accuracy exceeding expert models while still maintaining the interpretability offered by MLLMs. 
Extensive experiments confirm the superiority of DF-LLaVA, achieving both high accuracy and explainability in synthetic image detection.
\end{abstract}



\received{20 February 2007}
\received[revised]{12 March 2009}
\received[accepted]{5 June 2009}

\maketitle

\begin{figure}[ht]  
\centerline{  \includegraphics[width=\linewidth]{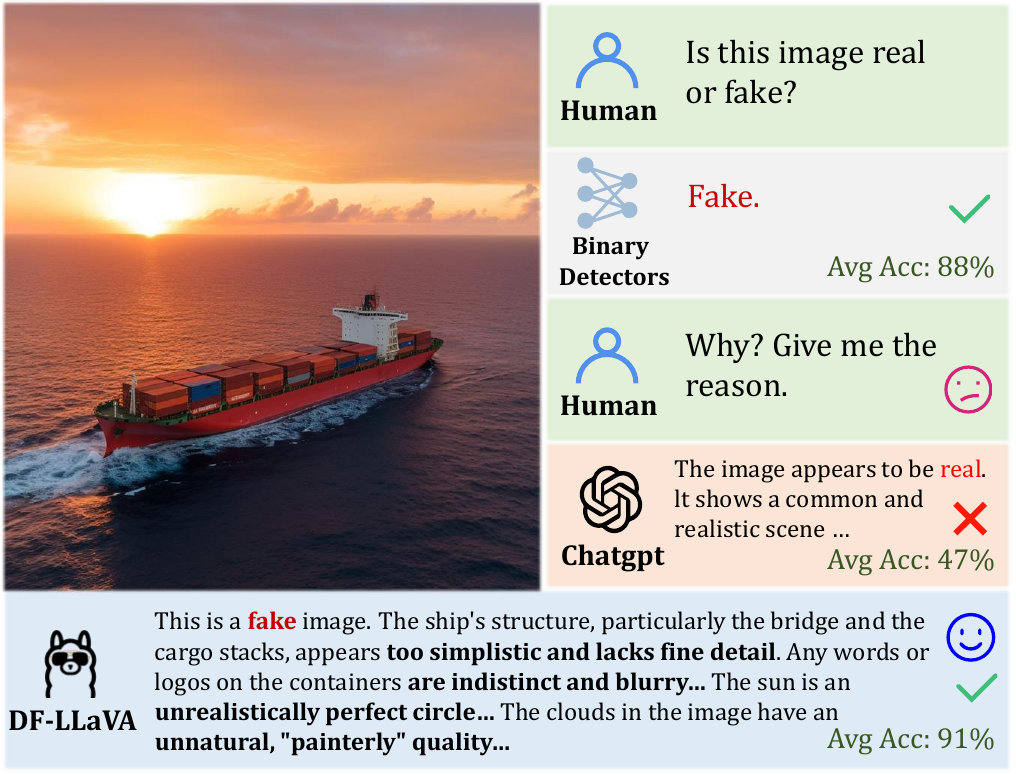}}
\caption{DF-LLaVA provides comprehensive artifact-level interpretability with detection accuracy outperforming expert models.}
\label{intro}
\end{figure}

\section{Introduction}
\label{sec:intro}
With the rapid progress of generative models, synthetic images have become increasingly prevalent across various domains. 
This surges in generated content raises growing concerns about image authenticity, making reliable detection and transparent explanation of potential forgeries an urgent task, with broad application demands in information security, digital forensics, and related fields.

Traditional detection methods typically formulate this problem as a binary classification of real versus fake images, providing a forgery probability or a classification result\cite{Haliassos2022Lips,li-2018,Jia-2024}. 
As shown in Fig \ref{intro}, existing synthetic image detection approaches still lack interpretability towards human users, \textbf{which is crucial for enabling human oversight, mitigating detection biases, and fostering greater confidence in the forensics of AI-generated content.}
This deficiency renders the detection results less accessible to the general public and undermines the perceived reliability of model outputs. 
Consequently, there is an urgent demand for developing expert models dedicated to fake image detection that achieve both high detection accuracy and transparency.

In response, a line of studies including \cite{Jia-2024,DBLP:journals/corr/abs-2503-14905}, Fakebench \cite{Li-2024}, and LOKI \cite{ye-2025} have explored the use of multimodal large language models(MLLMs).
Both closed-source GPT-4o \cite{gpt-4o}, Gemini \cite{Gemini2023}, Claude \cite{Anthropic2024Claude3} and open-source models InternVL \cite{chen-2024} deepseek \cite{lu-2024} Qwen2-VL \cite{Wang2024Qwen2VLEV} have been assessed. 
These models not only provide binary judgments of authenticity but also generate natural language rationales for their decisions, thereby improving the interpretability of detection outcomes (e.g., explaining that an image is likely synthetic due to irregular reflections in the eyes or inconsistent lighting across objects).
Nevertheless, the performance of these MLLMs in detection tasks remains inferior to that of domain-specific smaller expert models or human evaluators.

Recent studies have moved beyond general-purpose large models and proposed multimodal architectures tailored for synthetic image detection \cite{huang2024ffaa,chen2024x2,yan2025gpt,kang2025legion,ye-2025}. 
Representative methods such as FakeShield \cite{xu2025fakeshield} and ForgeryGPT \cite{liu2024forgerygpt} investigate the capability of large models to identify and localize forgery cues, while providing textual explanations for manipulated synthetic images. 
Nevertheless, the visual artifacts addressed in these works are predominantly associated with image manipulation, where inconsistencies typically appear in transitional regions and are characterized by salient boundary or edge artifacts. 
In contrast, our work focuses on images generated directly by synthesis models, which tend to exhibit structural, distortion, or physical artifacts.
Moreover, despite their reasoning and explanatory strengths, these models still fall short of expert detectors in terms of pure real-versus-fake classification performance.

To address the above challenges, we introduce DF-LLaVA, as shown in Fig \ref{intro}, a MLLM specifically designed for accurate synthetic image detection and artifact interpretation. 
We analyze prior works, reveal that the discriminative potential of MLLMs just lies within the vision encoder.
Building upon this observation, we introduce an auxiliary classifier to enable prompt-guided knowledge injection, effectively transferring discriminative cues from the frozen vision encoder into LLaVA's reasoning process.
Furthermore, our mechanism study reveals that a recurrent conflict between the predictions of the auxiliary classifier and the LLaVA backbone is the primary cause of erroneous final outputs.
To address this issue, we further propose a self-reflection mechanism during inference, which is driven by prediction conflict and refines the model’s final output, thereby improving detection precision and robustness.

As illustrated in Fig. \ref{teaser}, DF-LLaVA identifies artifacts produced by synthetic image models from diverse perspectives—structural, distortion, and physical—thereby enhancing interpretability towards human.  
Notably, DF-LLaVA achieves superior detection accuracy comparing with current expert models while offering strong interpretability of artifacts.  

Our main contributions are summarized as follows:
\begin{itemize}
\item We propose \textbf{DF-LLaVA}, which provides comprehensive artifact  interpretability with detection accuracy outperforming expert models.
\item We reveal that the discriminative potential of MLLMs lies within the vision encoder, and design a Prompt-Guided Knowledge Injection \textbf{(PGKI)} framework to leverage it.
\item We further introduce a novel Conflict-Driven Self-Reflection \textbf{(CDSR)} mechanism, enabling the model to re-evaluate and refine its predictions during inference.
\item Our method has been extensively evaluated on multiple benchmarks, achieving outstanding performance in both synthetic image detection and abnormal artifact explanation.
\end{itemize}

\section{Related Work}
\subsection{Synthetic Image Detection}
Synthetic image detection is commonly formulated as a binary classification task, aiming to distinguish real images from synthetic ones using data-driven approaches \cite{Jia-2024,li-2018,Zhao-2021,Haliassos2022Lips}. 
Early and mainstream methods are primarily built upon CNNs \cite{LeCun-1998} and Transformers \cite{Vaswani-2017}.
While these classifiers achieve strong in-distribution performance, later studies revealed that they struggle to generalize across generative models unseen during training \cite{Cozzolino-2018,Zhang-2019}, limiting their robustness in realistic scenarios.
Hence, the idea of learning classifiers that generalize to other generative models started gaining attention \cite{ojha-2023,Yan-2023}, especially under conditions where the training and testing domains are not aligned.
While these approaches achieve strong performance under controlled settings, they typically provide only binary authenticity predictions (real or fake) without offering explanations or insights into the underlying decision process.

Recently, several works including FakeBench \cite{Li-2024}, LOKI \cite{ye-2025}, and related studies \cite{Jia-2024}, demonstrate that MLLMs are capable of delivering detection results accompanied by natural-language explanations, thereby improving interpretability. 
Further efforts, such as Fakeshield \cite{xu2025fakeshield}, SIDA \cite{C16}, and ForgeryGPT \cite{liu2024forgerygpt} focus on detecting and explaining artifacts in manipulated synthetic data.   
However, these methods mainly target tampering artifacts, whereas our work focuses on directly synthesized images, where the dominant cues are structural distortions and warping artifacts rather than localized manipulations.

\subsection{Knowledge Injection}
Recent research on enhancing LLMs for domain-specific task performance has largely focused on knowledge injection techniques that adapt general pretrained models to specialized contexts. 
Knowledge injection can be broadly categorized into prompt-based methods, fine-tuning regimes, and retrieval-augmented generation.\\
\textbf{Prompt-Based method.} 
Prompt tuning inject domain knowledge by altering model input without modifying the core model weights.
Prompt enables LLMs to perform few-shot or zero-shot tasks through carefully designed input templates and auxiliary tokens that steer generation toward target behavior \cite{brown2020language,sahoo2024systematic,liu2023pre}.   \\
\textbf{Fine-Tuning.} 
Fine-tuning remains a core method for knowledge injection, modifying model parameters based on domain-specific datasets to internalize new expertise. 
Traditional full fine-tuning updates all weights but is increasingly replaced by Parameter-Efficient Fine-Tuning (PEFT) methods \cite{hu2022lora,li2021prefix,houlsby2019parameter,liu2022p} that adjust only a small fraction of the model’s parameters to reduce compute and memory costs while preserving general capabilities. 
Among the most influential PEFT techniques is LoRA (Low-Rank Adaptation) \cite{hu2022lora}, which reparameterizes weight updates using low-rank matrices to drastically shrink the number of trainable parameters. 
Subsequent variations including QLoRA\cite{dettmers2023qlora} extend LoRA by integrating quantization for further efficiency
and HydraLoRA\cite{tian2024hydralora} equips a shared backbone with multiple task-specific heads, enabling flexible multi-task adaptation.\\
\textbf{Retrieval-Augmented Generation (RAG).} 
RAG techniques inject domain knowledge at inference-time by coupling pretrained LLMs with external knowledge repositories \cite{lewis2020retrieval}.
The method has been shown to substantially improve factual accuracy and reduce hallucination, particularly in knowledge-intensive applications.
To further enhance structured and relational reasoning, knowledge graph–augmented RAG methods integrate graph-structured knowledge into the retrieval process, enabling explicit modeling of entity relations and domain hierarchies, as exemplified by GraphRAG and KAG \cite{edge2024local,liang2025kag}.

\begin{figure*}[!t]
\centerline{  \includegraphics[width=\linewidth]{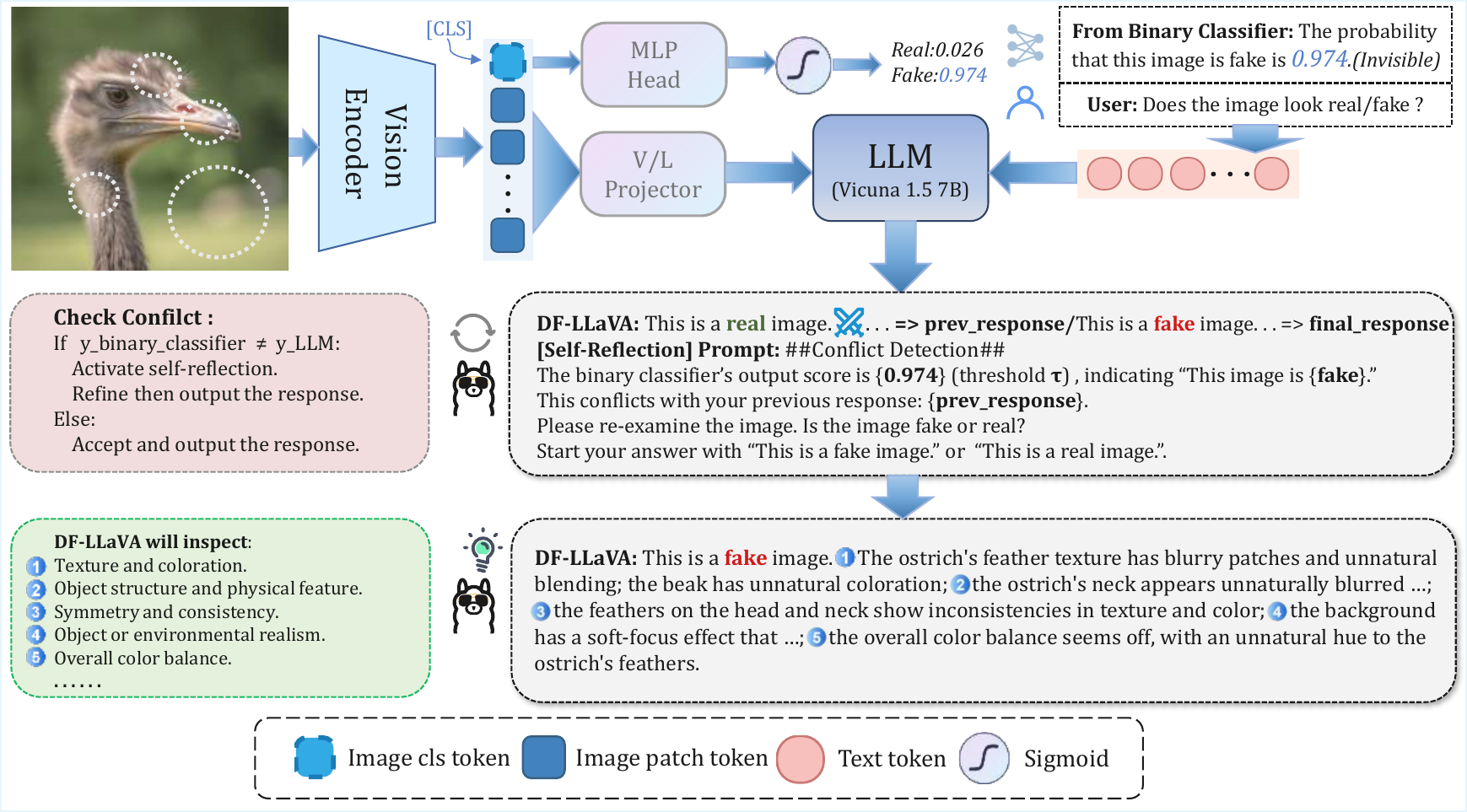}}
\caption{Overview of DF-LLaVA during inference. 
DF-LLaVA leverages its frozen vision encoder via a binary classifier for initial authenticity estimation.
The probabilistic output is used as reference in prompts, based on which DF-LLaVA makes its prediction. 
The prediction then undergoes a conflict check and a possible self-reflection process from model to ensure its precision and robustness.
Finally, artifacts are explained from various perspectives.
}
\label{teaser}
\end{figure*}

\subsection{Self-reflection}
LLMs could correct their own response using self-reflection while integrating external feedback from humans \cite{zheng2023progressive}, tools \cite{gou2023critic}, external metrics, or other LLMs \cite{paul2024refiner}. 
Another self-reflect strategy relies solely on the model’s internal reasoning capability \cite{pan2023}. 
Empirical studies show that these reflective mechanisms can lead to measurable improvements across a variety of tasks \cite{chen2023teaching}. 
Notably, self-reflection is not limited to any specific model architecture, observed across both proprietary models \cite{Jaech-2024} and open-source systems \cite{Guo-2025,Olmo-2024,Yang-2025} indicating that it may be a general emergent property of optimizing for complex reasoning objectives.

Recently, there are several works applying self-reflection to Vision Language Models(VLMs).
\cite{cheng-2024} proposes the R3V framework, which enables VLMs to iteratively reflect on chain-of-thought rationales in visual–language reasoning tasks, and guides the model to learn from its mistakes through a self-reflection loss.
VL-Rethinker \cite{wang-2025} addresses the insufficient reflection capability of VLMs, which enhances the model’s slow-thinking ability via reinforcement learning and a Forced Rethinking mechanism.
Qwen-Look-Again \cite{Chu-2025} analyzes the importance of reflection for visual attention in VLM reasoning, and proposes a mechanism that allows the model to re-attend to visual information during reasoning.
In our work, we design a self-reflection module which is activated when a conflict arises between the LLaVA's and the auxiliary model's predictions.

\begin{figure*}[!t]
\centerline{  \includegraphics[width=\linewidth]{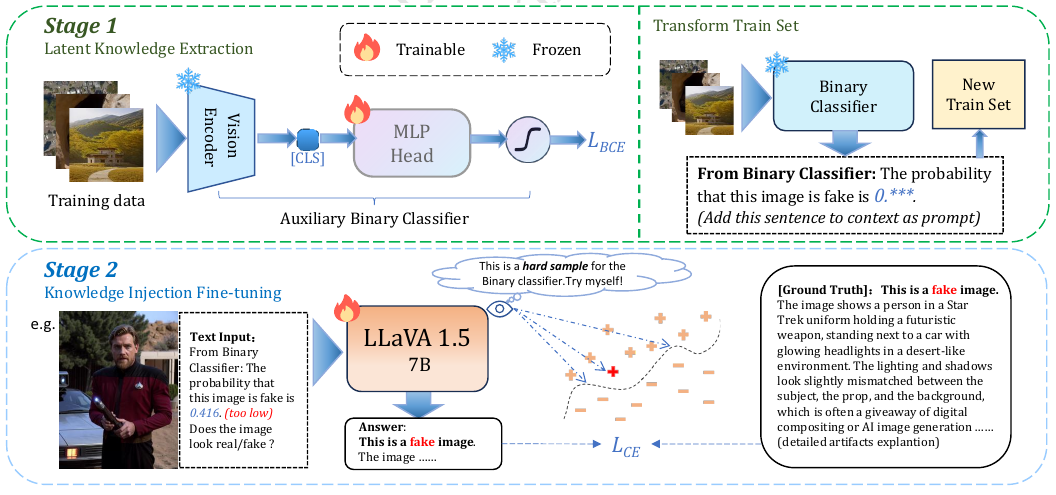}}
\caption{Overview of DF-LLaVA during training. 
(a) In Stage 1, we train an auxiliary binary classifier, whose predictions are injected into the train set as additional prompts. 
(b) In Stage 2, LLaVA is finetuned on this enriched dataset, learning to cooperate with the auxiliary classifier.}
\label{method}
\end{figure*}

\section{Preliminary}
Our framework is built upon the architecture of LLaVA-v1.5\cite{llava}, as illustrated in Fig. \ref{teaser}, which consists of four core components: i) a Vision Encoder, ii) a Vision/Language Projector, iii) a MLP Head iv) a Large Language Model (LLM). 
We detail each component as follows:  

\noindent\textbf{Vision Encoder}: We utilize the pretrained CLIP-ViT(L-14) \cite{C35Clip}'s visual branch as vision encoder. 
It produces 577 tokens for each image, including 576 patch tokens $V_{\text{patch}} \in \mathbb{R}^{N \times d_v}$ and a special \texttt{[CLS]} token $V_{\text{cls}} \in \mathbb{R}^{d_v}$ representing the global image feature:
\begin{equation}
     V_{\text{cls}}, V_{\text{patch}} = \text{CLIP-ViT}(I),\quad V_{\text{cls}} \in \mathbb{R}^{d_v},  V_{\text{patch}} \in \mathbb{R}^{N \times d_v} \
\end{equation}
\begin{equation}
    V = \text{CLIP-ViT}(I) \in \mathbb{R}^{N \times d_v}
\end{equation}
where $N = \frac{HW}{P^2}$ denotes the number of patches ($P=14$), and $d_v=1024$ the feature dimension. 

\noindent \textbf{Multi-modal Projector}: A two-layer MLP adapter bridges visual and textual modalities.
\begin{equation}
    \begin{aligned}
        H &= \text{GeLU}(V_{\text{patch}}W_1 + b_1) \\
        Z &= HW_2 + b_2
    \end{aligned}
\end{equation}
where $W_1 \in \mathbb{R}^{1024 \times 4096}$, $W_2 \in \mathbb{R}^{4096 \times 4096}$ are learnable parameters. The projected features $Z \in \mathbb{R}^{N \times 4096}$ combine with text embeddings of the task prompt $P$ through concatenation.

\noindent \textbf{MLP Head}: The lightweight MLP head takes $V_{\text{cls}}$ as input and produces a scalar probability via a Sigmoid function:
\begin{equation}
\begin{aligned}
    H_1 &= \text{LeakyReLU}(V_{\text{cls}}W_1 + b_1, \, 0.01) \\
    H_2 &= \text{Dropout}(H_1, \, p=0.3) \\
    \hat{y} &= \sigma(H_2 W_2 + b_2)
\end{aligned}
\end{equation}
where $W_1 \in \mathbb{R}^{1024 \times 10}$, $W_2 \in \mathbb{R}^{10 \times 1}$ are learnable parameters, $b_1$ and $b_2$ are biases, and $\sigma(\cdot)$ is the Sigmoid function.
\begin{equation}
    \sigma(x) = \frac{1}{1 + e^{-x}}
\end{equation}
During the first stage of PGKI, the MLP head is optimized using the Binary Cross-Entropy (BCE) loss:
\begin{equation}
    \mathcal{L}_{\text{BCE}} = - \big[ y \log \hat{y} + (1-y) \log (1-\hat{y}) \big]
\end{equation}
where $y \in \{0,1\}$ is the ground-truth label, and $\hat{y} \in [0,1]$ is the predicted probability.

\noindent \textbf{Large Language Model}: We utilize Vicuna-v1.5-7B\cite{vicuna} as our base LLM. 

\section{Methodology}
\label{sec:pagestyle}
\subsection{Latent Knowledge Extraction}
Recent studies \cite{Jia-2024, Li-2024, ye-2025} indicate that, despite their strong abilities in textual explanation, large pretrained multimodal models struggle to identify AI-generated images or distinguish manipulated ones within a collection. 
\cite{DBLP:journals/corr/abs-2503-14905} explores extracting visual representations from the final layer of LLaVA and evaluates their effectiveness for image authenticity detection using a linear probe.
The strategy achieves accuracy of roughly 70\%, suggesting the internal representations of large vision-language models inherently contain cues that differentiate real from synthetic images.
Similarly, UnivFD \cite{ojha-2023} demonstrates that CLIP’s visual feature space exhibits strong discriminative ability, where simple classifiers such as a linear probe or k-nearest neighbor achieve over 90\% accuracy across synthetic images generated by diverse methods. 
These findings inspire us to speculate that, during the propagation of image features from the vision encoder through the language model layers, the necessary information required for authenticity verification tends to be lost.

Therefore in this stage, our objective is to extract the latent discriminative knowledge embedded within the visual encoder and transfer it to the LLaVA itself, thereby achieving a win-win scenario of both high detection accuracy and improved interpretability.
As shown in Fig.\ref{method}, given that LLaVA’s visual encoder is derived from CLIP, we train a binary classifier on CLIP-ViT's [CLS] token on the train set using BCE loss in the initial step, serving as a way to extract knowledge. 
Since the classifier achieves relatively high accuracy, we treat its probabilistic outputs as embedded knowledge from the vision encoder. 
Subsequently, we freeze the binary classifier and directly inject these predictions into the prompts.  
By injecting the classifier’s output in textual form, the auxiliary knowledge is introduced through the same interface as natural language reasoning, making it directly accessible to the LLM’s decision process.
Consequently, we design a prompt template:
\begin{small}
\begin{verbatim}
From Binary Classifier: 
The probability that this image is fake is {P}.
\end{verbatim}
\end{small}
and use it to augment the entire training set.

Using probabilistic scores instead of hard labels preserves uncertainty information, allowing the model to reason about ambiguous cases rather than being overconfidently guided.
Compared with direct knowledge distillation \cite{Hinton2015DistillingTK}, this prompt-based approach does not limit the maximum achievable accuracy of the MLLM.
In other words, the MLLM can selectively accept or ignore the auxiliary predictions from the binary classifier by considering both the input images and the associated probabilistic scores, rather than being forced to align with the classifier’s outputs.

\subsection{Knowledge Injection Fine-tuning}
After obtaining the augmented training set, we freeze the vision encoder of LLaVA and perform full parameter fine-tuning on the multi-modal projector and LLM on the dataset. 
Freezing the vision encoder prevents degradation of its pretrained discriminative representations, while allowing the projector and LLM to adaptively learn how to utilize the injected knowledge.
During fine-tuning, LLaVA is gradually adapted to the classifier’s decision boundary and hard samples, learning to leverage the classifier as auxiliary guidance for detection.
In particular, samples near the classifier’s decision boundary provide informative supervision signals, encouraging the model to focus on subtle artifacts that are difficult to detect.
This fine-tuning strategy effectively aligns the language model’s reasoning with the vision encoder’s latent discriminative knowledge, without sacrificing generative flexibility.

\subsection{Conflict-Driven Self-Reflection}
Through mechanism analysis in \ref{Mechanism Analysis}, we found that the primary prediction errors of DF-LLaVA occur when the outputs of the main model conflict with those of the auxiliary binary classifier. 
Such conflicts indicate cases where the model’s internal reasoning diverges from strong visual priors, making them particularly informative for triggering corrective reasoning.
So we design and introduce an innovative conflict-driven self-reflection mechanism. 
Unlike unconditional self-reflection strategies, our conflict-driven design avoids unnecessary re-evaluation, reducing computational overhead while focusing reflection on high-risk samples.
Specifically, when the prediction from LLaVA disagrees with the auxiliary classifier, the model is prompted to reflect on its initial response through a self-reflection prompt template as follows:

\begin{small}
\begin{verbatim}
##Conflict Detection##
The binary classifier's output score is {S} (threshold {T}),
indicating “This image is {fake if S > T else real}.”
This conflicts with your previous response: {prev_response}.
Please re-examine the image.
Is the image fake or real?
Start your answer with "This is a fake image." 
or "This is a real image."
\end{verbatim}
\end{small}
This mechanism guides DF-LLaVA to re-evaluate and refine its first-round response, effectively allowing the model to reason over both the visual content and auxiliary knowledge. 
As a result, the final output integrates the complementary strengths of the main model and the binary classifier, improving detection accuracy and producing more reliable, explainable judgments.

\section{Experiments}
\label{sec:typestyle}
\subsection{Experiment Setting}
\textbf{Dataset.} We train the LLaVA on the entire FakeClue \cite{DBLP:journals/corr/abs-2503-14905} train set, with 10\% of the data split off as a validation set for the binary classifier. We then extensively evaluate our method on three commonly used benchmarks for synthetic image detection: FakeClue, LOKI \cite{ye-2025}, and DMImage \cite{C10}.
\textbf{FakeClue} covers 7 different categories of images.
, with over 100k image samples. 
It is organized as image-caption pairs for both the image and its artifact explanation in natural language. 
The test set contains 5k samples covering various image types.
\textbf{LOKI} is a recently proposed benchmark for evaluating MLLMs in general synthetic detection tasks.
Beyond just distinguishing real from fake, it also includes human manually annotated fine-grained image artifacts.
\textbf{DMimage} is a large-scale dataset designed to evaluate models in detecting synthetic images generated by diffusion models.

\begin{table*}[!t]
\centering
\begin{tabular}{rccccccccc}
\toprule[0.05em]
  & Method & \multicolumn{4}{c}{FakeClue} & \multicolumn{4}{c}{LOKI}  \\
   \cmidrule(r){3-6} \cmidrule(r){7-10} 
   && Acc $\uparrow$ & F1 $\uparrow$ & $\text{ROUGE}\_L \uparrow$& CSS $\uparrow$ & Acc $\uparrow$& F1 $\uparrow$ & $\text{ROUGE}\_L$ & CSS $\uparrow$ \\
   \hline
    & Deepseek-VL2-small\cite{lu-2024} & $40.4$ & $54.2$ & $17.1$ & $50.4$ & $25.3$ & $38.7$ & $16.4$ & $39.1$ \\
    & Deepseek-VL2\cite{lu-2024} & $47.5$ & $54.1$ & $17.2$ & $50.5$ & $43.1$ & $39.2$ & $16.9$ & $38.8$ \\
    & InternVL2-8B\cite{chen-2024}  & $50.6$ & $49.0$ & $18.0$ & $58.1$ & $52.6$ & $34.0$ & $17.9$ & $47.2$ \\
    & InternVL2-40B\cite{chen-2024}  & $50.7$ & $46.3$ & $17.6$ & $55.2$ & $50.7$ & $37.6$ & $\mathbf{18.4}$ & $47.3$ \\
    & Qwen2-VL-7B\cite{Wang2024Qwen2VLEV} & $45.7$ & $59.2$ & $26.6$ & $56.5$ & $57.1$ & $35.0$ & $18.2$ & $38.4$ \\
    & Qwen2-VL-72B\cite{Wang2024Qwen2VLEV} & $57.8$ & $56.5$ & $17.5$ & $54.4$ & $55.4$ & $40.9$ & $17.3$ & $43.2$ \\
    & GPT-4o(2024-08-06)\cite{gpt-4o} & $47.4$ & $42.0$ & $13.4$ & $40.7$ & $63.4$ & $57.2$ & $14.7$  & $35.4$ \\
    
    \midrule
    & (CVPR'20) CNNSpot \cite{C48} & $43.1$ & $9.8$ & - & - & $43.1$ & $11.4$ & - & - \\
    & (AAAI'24) FreqNet \cite{Tan-2024} & $48.7$ & $39.3$ & - & -  & $58.9$ & $50.6$ & - & - \\
    & (CVPR'24) Fatformer \cite{Liu-2024} & $54.5$ & $45.1$ & - & - & $58.8$ & $48.4$ & - & - \\
    & (CVPR'23) UnivFD \cite{ojha-2023} & $63.1$ & $46.8$ & - & - & $49.0$ & $35.8$ & - & - \\
    & (ICLR'25) AIDE$^{*}$ \cite{yan-2025} & $85.9$ & $94.5$ & - & - & $65.6$ & $80.2$ & - & - \\
    & (CVPR'24) NPR$^{*}$ \cite{Tan2-2024} & $90.2$ & $91.6$ & - & - & $77.4$ & $80.0$ & - & - \\
    
    \midrule
    & $\dagger$ ABC & $91.1$ & $93.2$ & - & - & $76.0$ & $80.8$ & -  & - \\

    \midrule
    & LLaVA-LoRA & $66.5$ & $ 70.9$ & $43.8$ & $82.7$ & $61.8$ & $65.1$ & $15.0$ & $47.5$ \\
    & LLaVA-FullFT & $90.3$ & $92.2$ & $54.0$ & $87.3$ & $73.4$ & $79.2$ & $15.2$ & $48.8$ \\

    & \textbf{DF-LLaVA(Ours)}  & $\mathbf{94.5}$ & $\mathbf{96.1}$ & $\mathbf{56.8}$ & $\mathbf{89.7}$ & $\mathbf{78.2}$ &  $\mathbf{83.7}$ & $16.3$ & $\mathbf{52.0}$ \\

\hline
\end{tabular}
\caption{
The experimental results evaluated on the FakeClue and LOKI datasets include both synthetic detection and artifact explanation performance. 
$\dagger$ ABC denotes our auxiliary binary classifier.
$*$ denotes methods trained on FakeClue.
}
\label{tab:main results}
\end{table*}

\noindent\textbf{ComparedBaselines.}
For tasks requiring both synthetic detection and artifact explanation(e.g.,
FakeClue, LOKI), we compare against a range of general-purpose MLLMs, including closed-source models such as GPT-4 and open-source counterparts like Qwen2-VL, InternVL2, and DeepSeek-VL2.
For pure synthetic data detection, we further compared with recent advanced expert methods.
Additionally, we introduce LLaVA-LoRA and LLaVA-FullFT as fine-tuning baselines to assess the effectiveness of parameter-efficient and full fine-tuning strategies, respectively.

\noindent\textbf{EvaluationMetrics.}
The evaluation metrics are categorized into two tasks: detection and artifact explanation. 
Detection accuracy is evaluated with accuracy (Acc) and F1 scores, while artifact explanation accuracy is measured using CSS and ROUGE\_L.

\noindent\textbf{Implementation Details.}
We conduct experiments on LLaVA-1.5 7B\cite{llava} and perform full fine-tuning on the LLM and V/L projector. 
We train our models and baseline method
LLaVA-LoRA, LLaVA-FullFT on four A6000 48G GPUs with a batch size of 32.
Regarding hyperparameter settings, we adhere to most settings of LLaVA, except for using a maximum learning rate of 1e-4 and training for 2 epochs.
For auxiliary binary classifier, we employ the [CLS] feature from the same layer as the patch features fed into LLaVA’s V/L projector.
And we set the classifier’s decision threshold to 0.58 consistently across all three benchmarks.
Detailed training configurations will be provided in the accompanying code repository.

\begin{table}[!ht]
\centering
\resizebox{\columnwidth}{!}{
\begin{tabular}{ccccccc}
\cmidrule[0.08em](lr){1-7}
\multirow{2}{*}{Method} & \multicolumn{2}{c}{Real} & \multicolumn{2}{c}{Fake} & \multicolumn{2}{c}{Overall} \\ 
\cmidrule(r){2-3} \cmidrule(r){4-5} \cmidrule(r){6-7}
 & Acc & F1 & Acc & F1 & Acc & F1 \\ 
\cmidrule[0.05em](lr){1-7}
(CVPR'20) CNNSpot\cite{C48}      &  87.8   & 88.4   &  28.4  & 44.2  &  40.6  &  43.3   \\
(CVPR'20) GramNet\cite{C30}     & 62.8    & 54.1   & 78.8  & 88.1  &  67.4  &  79.4   \\
(ICIP'22) Fusing\cite{C20}         &  87.7   & 86.1    & 15.5    & 27.2  &  40.4  &  36.5   \\
(ECCV'22) LNP\cite{C3}            & 63.1    & 67.4   & 56.9   & 72.5  &  58.2  &  68.3   \\
(CVPR'23) UnivFD \cite{ojha-2023}       & 89.4    & 88.3   & 44.9   & 61.2  &  53.9  &  60.7   \\
(ICLR'24) Antifake\cite{C4} &  91.3    & 92.5    & 89.3    & 91.2  &  90.6  &  91.2   \\
(CVPR'25) SIDA\cite{C16} &  \textbf{92.9} & 93.1 & 90.7 & 91.0 & \textbf{91.8} & 92.4 \\
\cmidrule[0.05em](lr){1-7}
$\dagger$ ABC       & 88.9    & \textbf{94.6}   & 90.7   & 95.2  &  90.6  &  93.6   \\
\cmidrule[0.05em](lr){1-7}
LLaVA-LoRA          & 71.1    & 83.1   & 66.8   &  80.1  &  67.8  &  76.0 \\ 
LLaVA-FullFT          & 70.9    & 74.2   & 92.6   &  94.5  &  85.9  &  91.1 \\ 
\textbf{DF-LLaVA(Ours)}   & 85.3    & 88.9   & \textbf{95.7}   &  \textbf{97.9}  &  91.7  &  \textbf{94.6} \\
\cmidrule[0.08em](lr){1-7}
\end{tabular}
}
\caption{
Comparison with other detection methods on the DMimage\cite{C10} dataset.
$\dagger$ ABC denotes our auxiliary binary classifier.}
\label{table:dmimage}
\end{table}

\subsection{Experimental Results}
Our primary experimental results are summarized in Table \ref{tab:main results}. 
We compare DF-LLaVA with fine-tuning baselines, leading expert models and general-purpose MLLMs, demonstrating its superior performance across multiple metrics, excelling in both synthetic image detection and artifact interpretation.
Specifically, compared to the powerful open-source model Qwen2-VL-72B, DF-LLaVA achieves an average improvement of 29.8\% in Acc and 41.2\% in F1 on both FakeClue and LOKI.
Additionally, compared to the leading expert model NPR or AIDE, which is also trained on FakeClue, DF-LLaVA also achieves an average improvement of 6.6\% in Acc and 3.3\% in F1.
Furthermore, DF-LLaVA surpasses the baseline methods substantially, including LLaVA-LoRA, LLaVA-FullFT and the auxiliary classifier in both accuracy and F1 score, indicating that it effectively exploits latent knowledge beyond explicit classifier signals and thereby highlighting the effectiveness of prompt-guided knowledge injection.
As reflected by the CSS and ROUGE\_L metrics, DF-LLaVA provides stronger interpretability than general MLLMs, offering more accurate and reliable artifact explanations.

\enlargethispage{-\baselineskip}
\begin{figure*}[!t]
\centerline{  \includegraphics[width=\linewidth]{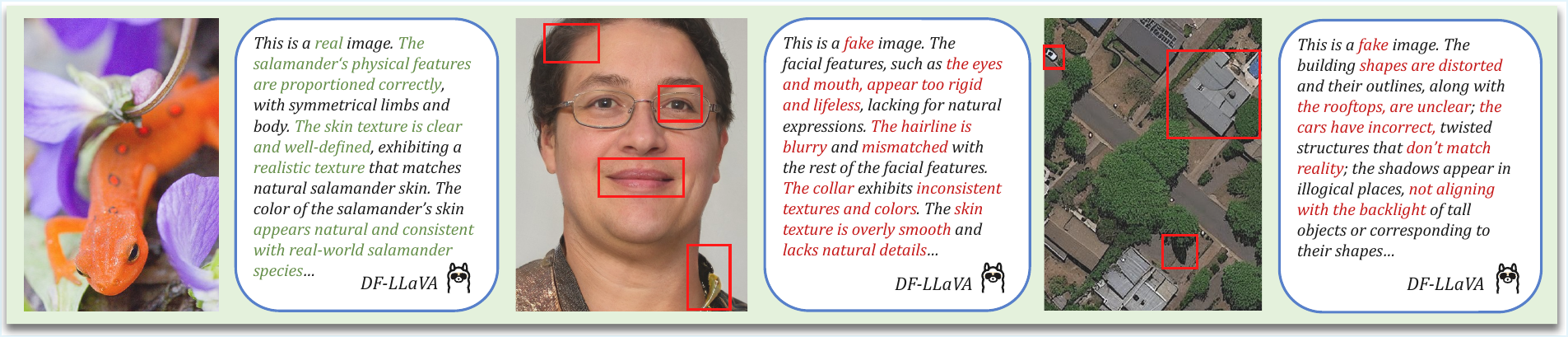}}
\caption{Qualitative evaluation.
DF-LLaVA can explain artifacts from structural, distortion, and physical perspectives, demonstrating superior detection and interpretability.}
\label{qua}
\end{figure*}

\begin{figure*}[htbp]
\centering

\begin{subfigure}[t]{0.33\textwidth}
\includegraphics[width=\textwidth]{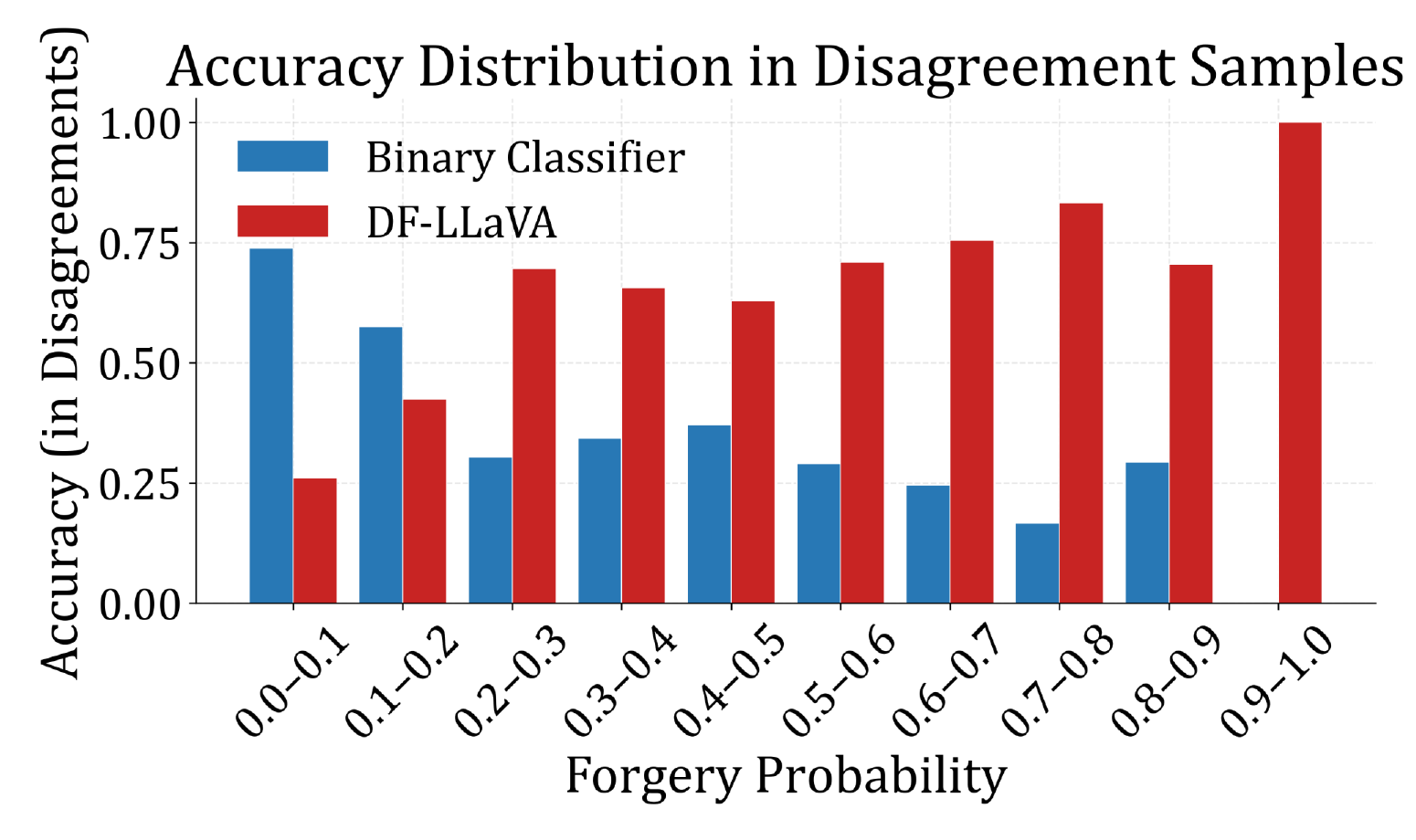}
\caption{Accuracy Distribution on FakeClue dataset}
\label{fig:1c}
\end{subfigure}
\hfill
\begin{subfigure}[t]{0.33\textwidth}
\includegraphics[width=\textwidth]{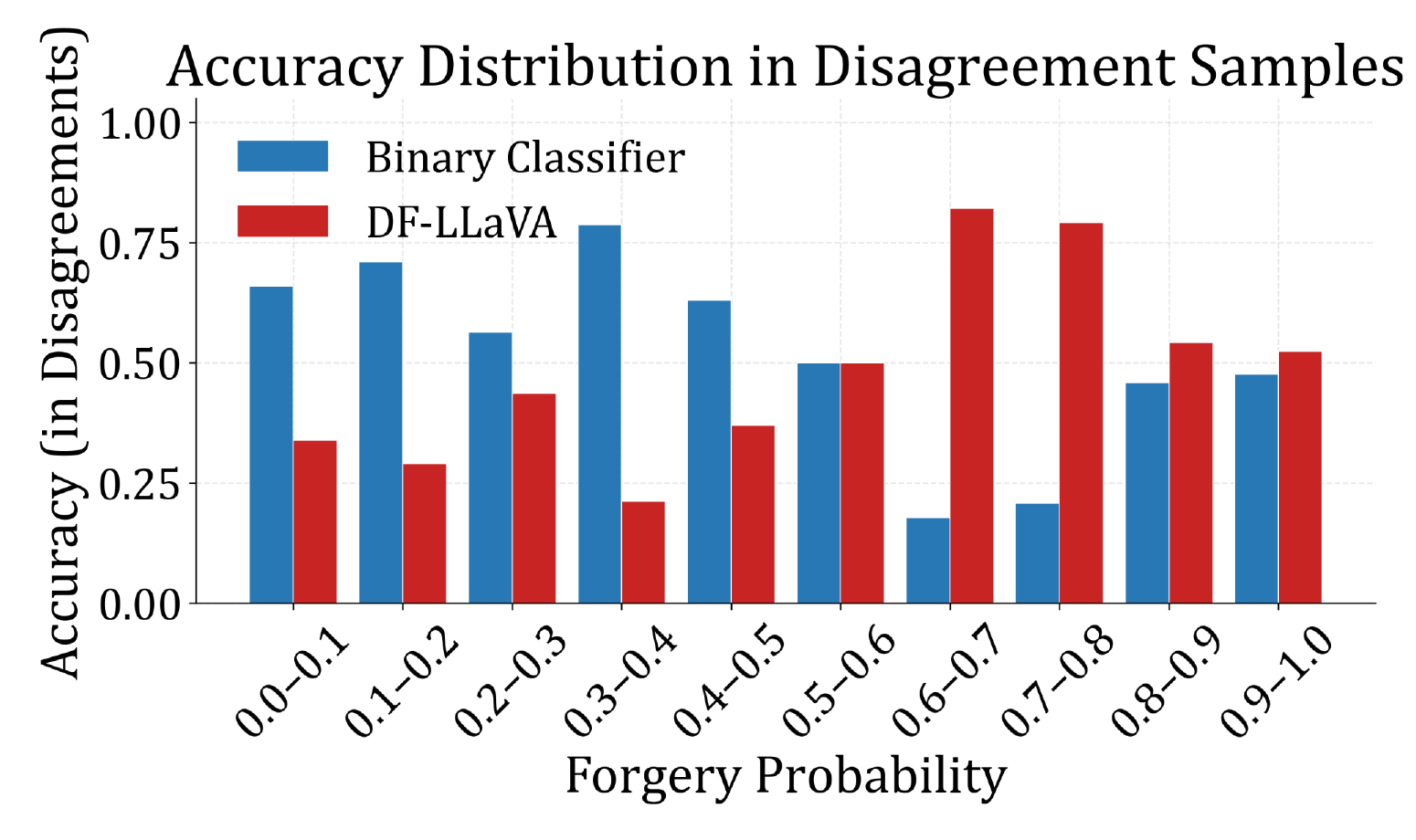}
\caption{Accuracy Distribution on LOKI dataset}
\label{fig:1e}
\end{subfigure}
\hfill
\begin{subfigure}[t]{0.33\textwidth}
\includegraphics[width=\textwidth]{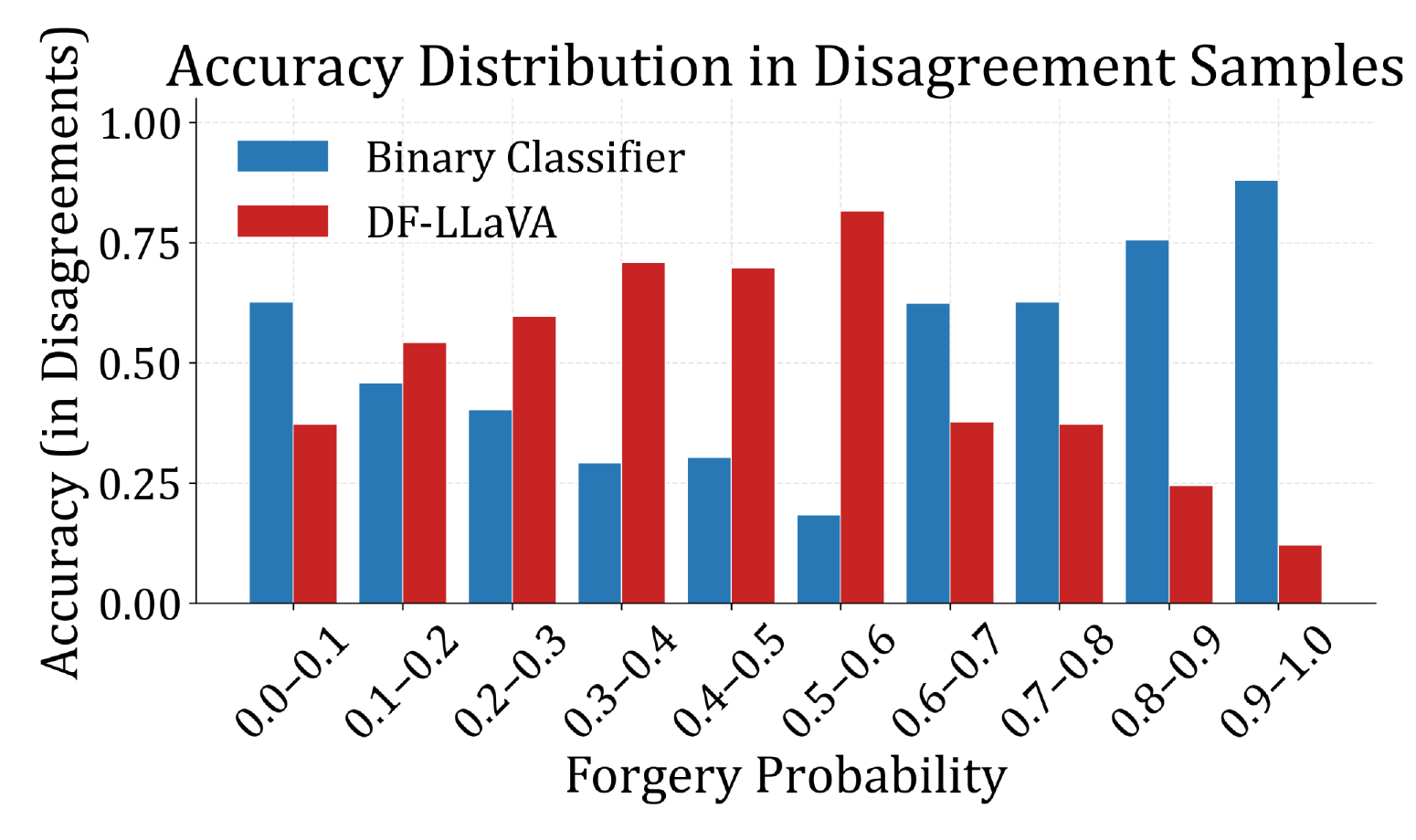}
\caption{Accuracy Distribution on DMimage dataset}
\label{fig:1a}
\end{subfigure}

\vspace{1em} 


\begin{subfigure}[t]{0.33\textwidth}
\includegraphics[width=\textwidth]{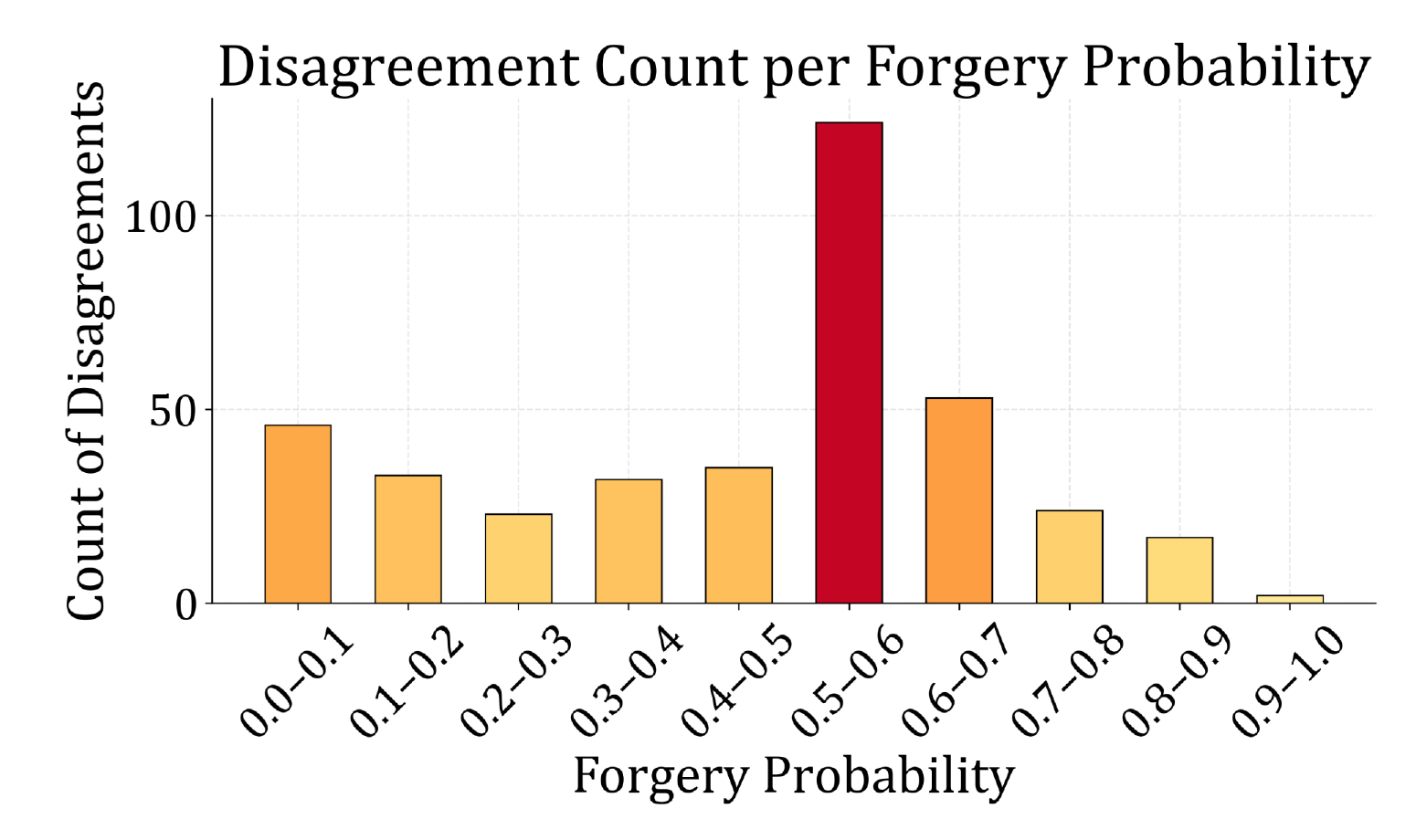}
\caption{Disagreement Count on FakeClue dataset}
\label{fig:1d}
\end{subfigure}
\hfill
\begin{subfigure}[t]{0.33\textwidth}
\includegraphics[width=\textwidth]{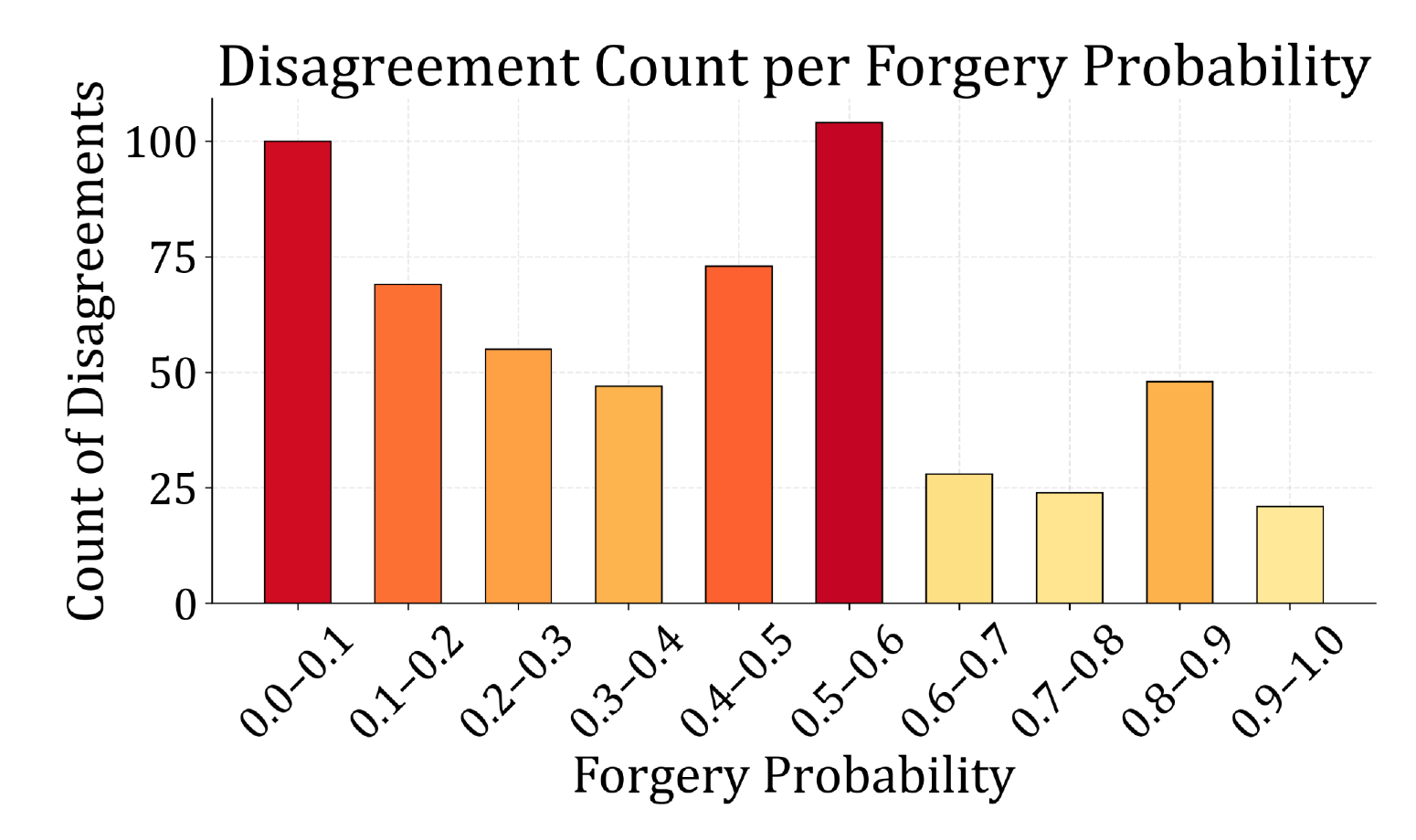}
\caption{Disagreement Count on LOKI dataset}
\label{fig:1f}
\end{subfigure}
\hfill
\begin{subfigure}[t]{0.33\textwidth}
\includegraphics[width=\textwidth]{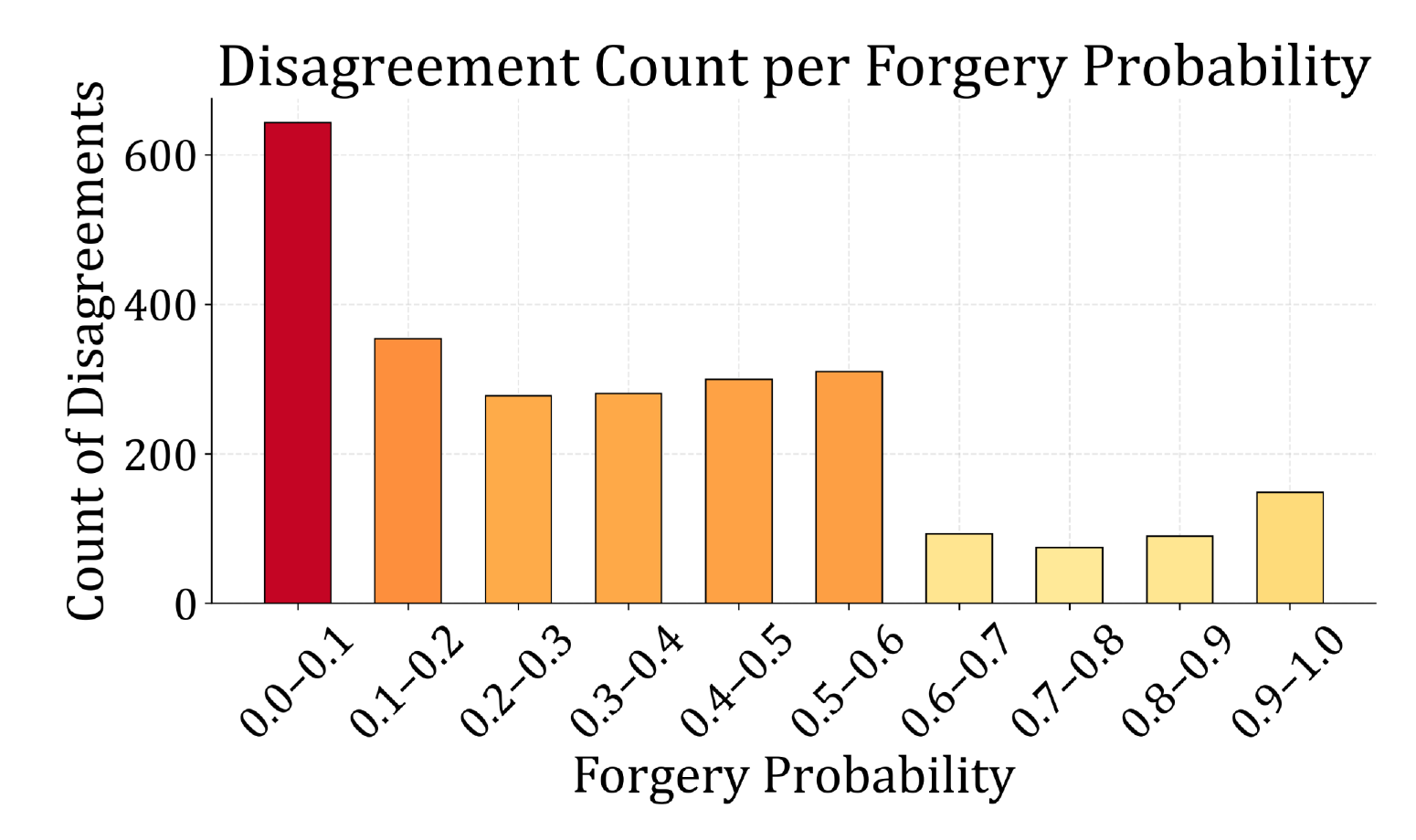}
\caption{Disagreement Count on DMimage dataset}
\label{fig:1b}
\end{subfigure}

\caption {Comparative visualization of discriminative performance and disagreement counts across different intervals between the binary classifier and DF-LLaVA on three benchmark datasets. The top row shows accuracy distributions; the bottom row shows disagreement counts. The visualization highlights DF-LLaVA's superior performance in ambiguous regions while revealing its tendency toward over-reasoning on high-confidence true samples.}
\label{fig:decision_boundaries_comparison}
\end{figure*}

We also present the generalization experiment results of DF-LLaVA on DMimage\cite{C10} in Table \ref{table:dmimage}, following Huang et al. \cite{C16}.
The experimental results demonstrate that DF-LLaVA achieves performance comparable to, or even surpassing expert models while retaining its capability for artifact explanation. 
In particular, regarding the overall F1 score and performance on the fake class, DF-LLaVA attains the best results, outperforming all expert models. 
Although DF-LLaVA performs relatively weaker than expert models on the real images—likely due to class imbalance in the training data, it still achieves substantial improvements over prior approaches, particularly in detecting fake images, which is the key objective in synthetic image detection. 

Fig.\ref{qua} presents DF-LLaVA’s qualitative evaluation. 
DF-LLaVA is capable of detecting synthesis-related defects—such as visual artifacts, abnormal textures, and structural inconsistencies—while providing fine-grained natural language rationales.
In contrast to approaches that rely on fixed probability thresholds, DF-LLaVA offers intuitive and interpretable explanations, which substantially improve detection transparency and support more trustworthy and reliable assessments of synthetic content.

\subsection{Mechanism Analysis}
\label{Mechanism Analysis}
As shown in Fig.\ref{fig:decision_boundaries_comparison}, to elucidate the mechanism underlying the cooperation between DF-LLaVA and its auxiliary classifier, we conduct a fine-grained analysis focusing on \emph{disagreement samples}, defined as instances in which the two models assign conflicting labels.
We then exclusively evaluate detection performance on this subset.

We observe a clear performance gap emerging within the uncertain probability interval from approximately 0.3 to 0.6.
Within this range, the auxiliary classifier exhibits substantial performance degradation; e.g., its accuracy drops to about 25\% on conflicting samples from FakeClue.
In contrast, DF-LLaVA relatively maintains a high level of accuracy in this interval. 
This result suggests that DF-LLaVA can effectively identify hard samples that challenge the auxiliary classifier, thereby enabling correct predictions even when the auxiliary classifier’s predictions are ambiguous.

On DMimage and LOKI, we observe that DF-LLaVA tends to question samples that are predicted as real, with prediction disagreements between DF-LLaVA and the auxiliary classifier predominantly concentrated in the high-confidence authentic interval ($0.0$–$0.1$).
In contrast, on FakeClue, the dataset used to train DF-LLaVA, the disagreements between the two models are more reasonably distributed and mainly cluster within the uncertain interval ($0.5$–$0.6$), reflecting normal decision behavior under an in-distribution setting.
We conclude that DF-LLaVA implicitly distinguishes between in-distribution and out-of-distribution samples, exhibiting normal uncertainty-aware behavior on training distribution data while adopting a more cautious and corrective decision strategy under distribution shift.

These observations highlight that model disagreements in critical probability intervals—whether arising from overconfident predictions on authentic samples or under-confidence on difficult cases. 
This motivates the introduction of the Conflict-Driven Self-Reflection mechanism, which leverages such conflicts to guide the model in re-evaluating challenging samples. 
Therefore, DF-LLaVA can better integrate information from both the main model and the auxiliary classifier, enhancing robustness and reliability under distribution shift.

\begin{table}[ht]
\centering
\setlength{\tabcolsep}{2.00mm}
\resizebox{\columnwidth}{!}{
\begin{tabular}{c c c c c c c}
\toprule
\multirow{2}{*}{\textbf{Setting}} 
& \multicolumn{2}{c}{\textbf{FakeClue}} 
& \multicolumn{2}{c}{\textbf{LOKI}}
& \multicolumn{2}{c}{\textbf{DMimage}} \\
\cmidrule(lr{0pt}){2-3}
\cmidrule(lr{0pt}){4-5}
\cmidrule(lr{0pt}){6-7}
& \textbf{\textit{Acc}} & \textbf{\textit{F1}} 
& \textbf{\textit{Acc}} & \textbf{\textit{F1}} 
& \textbf{\textit{Acc}} & \textbf{\textit{F1}} \\
\midrule


\rowcolor{gray!15}
LLaVA-FullFT 
& 90.3 & 92.2 
& 73.4 & 79.2
& 85.9 & 91.1 \\

+PGKI (w/o FT)
& 91.9{\textcolor{ForestGreen}{(+1.6)}} 
& 93.9{\textcolor{ForestGreen}{(+1.7)}} 
& 73.6{\textcolor{ForestGreen}{(+0.2)}} 
& 79.8{\textcolor{ForestGreen}{(+0.6)}}
& 86.3{\textcolor{ForestGreen}{(+0.4)}}
& 88.9{\textcolor{red}{(-2.2)}} \\

+PGKI 
& 93.4{\textcolor{ForestGreen}{(+3.1)}} 
& 95.0{\textcolor{ForestGreen}{(+2.8)}} 
& 76.7{\textcolor{ForestGreen}{(+3.3)}} 
& 82.5{\textcolor{ForestGreen}{(+3.3)}}
& 91.0{\textcolor{ForestGreen}{(+5.1)}}
& 94.1{\textcolor{ForestGreen}{(+3.0)}} \\

+PGKI +SR (w/o CD)
& 91.5{\textcolor{ForestGreen}{(+1.2)}} 
& 93.7{\textcolor{ForestGreen}{(+1.5)}} 
& 77.2{\textcolor{ForestGreen}{(+3.8)}} 
& 82.8{\textcolor{ForestGreen}{(+3.6)}}
& 91.2{\textcolor{ForestGreen}{(+5.3)}}
& 94.3{\textcolor{ForestGreen}{(+3.2)}} \\

+PGKI +CDSR
& \textbf{94.5}{\textcolor{ForestGreen}{(+4.2)}} 
& \textbf{96.1}{\textcolor{ForestGreen}{(+3.9)}} 
& \textbf{78.2}{\textcolor{ForestGreen}{(+4.8)}} 
& \textbf{83.7}{\textcolor{ForestGreen}{(+4.5)}}
& \textbf{91.7}{\textcolor{ForestGreen}{(+5.8)}}  
& \textbf{94.6}{\textcolor{ForestGreen}{(+3.5)}} \\
\bottomrule
\end{tabular}
}
\caption{Ablation studies on DF-LLaVA.
PGKI denotes the Prompt-Guided Knowledge Injection, and
CDSR denotes the Conflict-Driven Self-Reflection.
SR and CDSR are applied on top of the PGKI-enhanced model.}
\label{ablation}
\end{table}

\subsection{Ablation Study}
In this section, we present comprehensive ablation experiments on DF-LLaVA.
The results are summarized in Table \ref{ablation}.
We mainly evaluate the impact under the LLaVA-FullFT baseline setting. \\
\textbf{Influence of Prompt-Guided Knowledge Injection.} 
The results of our experiments were obtained using three different random seeds and are reported as their average.
When PGKI is applied without fine-tuning (w/o FT), i.e., directly injecting the auxiliary classifier’s prediction into the prompt for LLaVA’s reference, it brings only marginal improvement and even causes a drop in DMimage’s F1 score.
By contrast, applying PGKI with fine-tuning allows the model to use the auxiliary classifier head as a medium to utilize knowledge from the frozen vision encoder while simultaneously adapting to the classifier’s decision boundary. 
This results in substantial performance gains across all benchmarks.\\
\textbf{Self-Reflection.} 
The subsequent SR and CDSR experiments are conducted on top of the PGKI-enhanced model.
We get three extra variants of the self-reflection prompt using chatgpt, and then conducted experiments based on these four prompt templates, with the final results averaged.
Specifically, SR (w/o CD) performs self-reflection regardless of whether there is a conflict between DF-LLaVA and the auxiliary classifier’s predictions. 
While this introduces the self-reflection mechanism, the results are unstable, showing both slight improvements and drops compared to PGKI alone.
By incorporating conflict detection as a signal (CDSR), the model can more effectively identify inconsistencies and correct errors, leading to more accurate predictions. 
Combined with PGKI, this results in the most significant performance gains, yielding the full DF-LLaVA model.

\section{Conclusion}
In this paper, we presented DF-LLaVA, a novel and effective framework that unlocks the intrinsic discrimination potential of MLLMs for synthetic image detection. 
By employing prompt-guided knowledge injection, our framework successfully leverages knowledge inherent to the MLLM.
With the self-reflection mechanism during inference, DF-LLaVA can further refine its response, achieving strong detection accuracy comparable to or even exceeding expert models while preserving interpretability towards human.
In the future, we aim to develop a more unified and adaptive reasoning framework that enables MLLMs to dynamically adjust their discriminative behavior for synthetic image detection across diverse architectures and generation paradigms.

\begin{acks}
This work is supported by the National Natural Science Foundation of China (NO. 62572193), China Postdoctoral Science Foundation (NO. 2024M760930), the Open Research Fund of the Key Laboratory of Advanced Theory and Application in Statistics and Data Science, Ministry of Education, and the Fundamental Research Funds for the Central Universities.
\end{acks}

\vfill
\pagebreak

\bibliographystyle{ACM-Reference-Format}
\bibliography{ref_arxiv}


\end{document}